\documentclass[runningheads]{llncs}

\usepackage{indentfirst}
\usepackage{graphicx}
\usepackage{amsmath,amssymb} 
\usepackage{color}
\usepackage{multirow}
\usepackage{caption}
\usepackage{booktabs}
\usepackage{cite}
\usepackage{enumitem}
\usepackage{mmstyle}

\newcommand{\conv}{\circledast}
\makeatletter
\newcommand{\printfnsymbol}[1]{%
  \textsuperscript{\@fnsymbol{#1}}%
}
\makeatother

\begin{document}
\title{Rethinking the Form of Latent States \\ in Image Captioning} 

\titlerunning{Rethinking the Form of Latent States in Image Captioning}
%
\author{Bo Dai\thanks{equal contribution}~$^1$ \and
Deming Ye\printfnsymbol{1}~$^2$ \and
Dahua Lin~$^1$}
%
\authorrunning{B.Dai, D.Ye, and D.Lin}
%

\institute{CUHK-SenseTime Joint Lab, The Chinese University of Hong Kong \and
Department of Computer Science and Technology, Tsinghua University  \\
\email{db014@ie.cuhk.edu.hk}~~~
\email{ydm18@mails.tsinghua.edu.cn}~~~ 
\email{dhlin@ie.cuhk.edu.hk}}
\maketitle              


\begin{abstract}
RNNs and their variants have been widely adopted for image captioning.
In RNNs, the production of a caption is driven by a sequence of
latent states.
Existing captioning models usually represent latent states as vectors,
taking this practice for granted.
We rethink this choice and study an alternative formulation,
namely using two-dimensional maps to encode latent states.
This is motivated by the curiosity about a question:
\emph{how the spatial structures in the latent states affect
the resultant captions?}
Our study on
MSCOCO and Flickr30k
leads to two significant observations.
First, the formulation with 2D states is generally more effective in captioning,
consistently achieving higher performance with comparable parameter sizes.
Second, 2D states preserve spatial locality.
Taking advantage of this, we \emph{visually}
reveal the internal dynamics in the process of caption generation,
as well as the connections between input visual domain and output linguistic domain.
\end{abstract}


\section{Introduction}
\label{sec:intro}


Image captioning, a task of generating short descriptions for given images,
has received increasing attention in recent years.
Latest works on this task~\cite{vinyals2015show, xu2015show, rennie2016self, lu2016knowing}
mostly adopt the encoder-decoder paradigm,
where a recurrent neural network (RNN) or one of its variants,
\eg~GRU~\cite{cho2014learning} and LSTM~\cite{hochreiter1997long},
is used for generating the captions.
Specifically, the RNN maintains a series of \emph{latent states}.
At each step, it takes the visual features together with the preceding word as input,
updates the latent state,
then estimates the conditional probability of the next word.
Here, the latent states serve as pivots that connect between
the visual and the linguistic domains.

%
Following the standard practice in language models~\cite{cho2014learning, graves2013generating},
existing captioning models usually formulate the latent states
as \emph{vectors} and the connections between them as
fully-connected transforms.
Whereas this is a natural choice for purely linguistic tasks,
it becomes a question when the visual domain comes into play,
\eg~in the task of image captioning.

%
Along with the rise of deep learning,
convolutional neural networks (CNN) have become the dominant
models for many computer vision tasks~\cite{He2015, ILSVRC15}.
\emph{Convolution} has a distinctive property, namely \emph{spatial locality},
\ie~each output element corresponds to a local region in the input.
This property allows the spatial structures to be maintained
by the feature maps across layers.
The significance of spatial locality for vision tasks have been repeatedly
demonstrated in previous work~\cite{He2015, bau2017network, huang2015bidirectional, romera2016recurrent, li2017videolstm}.

%
Image captioning is a task that needs to bridge both the linguistic and the visual domains.
Thus for this task, it is important to capture and preserve properties of the visual content in the latent states.
This motivates us to explore an
alternative formulation for image captioning, namely
representing the latent states with 2D maps and connecting them via
convolutions.
As opposed to the standard formulation, this variant is capable of
preserving spatial locality, and therefore it may strengthen the role
of visual structures in the process of caption generation.

%
We compared both formulations, namely
the standard one with vector states and the alternative one that
uses 2D states, which we refer to as \emph{RNN-2DS}.
Our study shows:
(1) The spatial structures significantly impact the captioning
process. Editing the latent states, \eg~suppressing certain regions in the
states, can lead to substantially different captions.
(2) Preserving the spatial structures in the latent states is beneficial
for captioning.
On two public datasets, MSCOCO \cite{lin2014microsoft}~and Flickr30k \cite{young2014image},
RNN-2DS achieves notable performance gain consistently across different settings.
In particular, a simple RNN-2DS without gating functions already outperforms
more sophisticated networks with vector states, \eg~LSTM.
Using 2D states in combination with more advanced cells, \eg~GRU, can
further boost the performance.
(3) Using 2D states makes the captioning process amenable to
visual interpretation. Specifically, we take advantage of the spatial
locality and develop a simple yet effective way to identify the
connections between latent states and visual regions.
This enables us to visualize the dynamics of the states as a caption
is being generated,
as well as the connections between the visual domain and the linguistic domain.

%
In summary, our contributions mainly lie in three aspects.
First, we rethink the form of latent states in image captioning models, for which existing work simply
follows the standard practice and adopts the vectorized representations.
To our best knowledge, this is the first study that systematically explores
two dimensional states in the context of image captioning.
Second, our study challenges the prevalent practice,
which reveals the significance of spatial locality in image captioning
and suggests that the formulation with 2D states and convolution
is more effective.
Third, leveraging the spatial locality of the alternative formulation,
we develop a simple method that can visualize the dynamics of the latent states
in the decoding process.


\section{Related Work}
\label{sec:relwork}

\paragraph{\bf Image Captioning.}
Image captioning has been an active research topic in computer vision.
Early techniques mainly rely on detection results.
Kulkarni \etal~\cite{kulkarni2013babytalk} proposed to first detect visual
concepts including objects and visual relationships \cite{dai2017detecting}, 
and then generate captions by filling sentence templates.
Farhadi \etal~\cite{farhadi2010every} proposed to generate captions
for a given image by retrieving from training captions based on
detected concepts.

In recent years, the methods based on neural networks are gaining ground.
Particularly, the encoder-decoder paradigm~\cite{vinyals2015show}, which
uses a CNN~\cite{simonyan2014very} to encode visual features and then uses
an LSTM net~\cite{hochreiter1997long} to decode them into a caption, was shown
to outperform classical techniques and has been widely adopted.
Along with this direction,
many variants have been proposed \cite{xu2015show, dai2017towards, dai2017contrastive, yao2016boosting},
where 
Xu \etal~\cite{xu2015show} proposed to use a dynamic attention map to guide
the decoding process.
And Yao \etal~\cite{yao2016boosting} additionally incorporate visual
attributes detected from the images, obtaining further improvement.
While achieving significant progress,
all these methods rely on \emph{vectors}
to encode visual features and to represent latent states.

\paragraph{\bf Multi-dimensional RNN.}

Existing works that aim at extending RNN to more dimensions
roughly fall into three categories:

(1) RNNs are applied on \emph{multi-dimensional grids},
\eg~the 2D grid of pixels, via recurrent connections
along different dimensions~\cite{graves2007multi, zuo2015convolutional}.
Such extensions have been used in
image generation~\cite{wu2016deep} and CAPTCHA recognition~\cite{rui2013novel}.

(2) Latent states of RNN cells are stacked
across multiple steps to form feature maps.
This formulation is usually used to capture temporal statistics,
\eg~those in language processing \cite{wang2017hybrid, fu2017crnn}
and audio processing \cite{keren2016convolutional}.
For both categories above, the latent states are still represented
by \emph{1D vectors}. Hence, they are essentially different from this work.

(3) Latent states themselves are represented as multi-dimensional
arrays.
The RNN-2DS studied in this paper belongs to the third category,
where latent states are represented as 2D feature maps.
The idea of extending RNN with 2D states has been explored in various
vision problems, such as
rainfall prediction~\cite{xingjian2015convolutional},
super-resolution~\cite{huang2015bidirectional},
instance segmentation~\cite{romera2016recurrent},
and action recognition~\cite{li2017videolstm}.
It is worth noting that all these works focused on tackling visual tasks,
where both the inputs and the outputs are in 2D forms.
To our best knowledge, this is the first work that studies recurrent networks
with 2D states in image captioning.
A key contribution of this work
is that it reveals the significance of 2D states in connecting the visual
and the linguistic domains.

\paragraph{\bf Interpretation.}

There are studies to analyze recurrent networks.
Karpathy \etal~\cite{karpathy2015visualizing} try to interpret
the latent states of conventional LSTM models for natural language
understanding.
Similar studies have been conducted by Ding \etal~\cite{ding2017visualizing}
for neural machine translation.
However, these studies focused on linguistic analysis,
while our study tries to identify the connections between linguistic
and visual domains by leveraging the spatial locality of the
2D states.

Our visualization method on 2D latent states also differs from 
the attention module~\cite{xu2015show} fundamentally,
in both theory and implementation.
(1) Attention is a \emph{mechanism} specifically designed to guide the focus of a model,
while the 2D states are a form of \emph{representation}.
(2) Attention is usually implemented as a sub-network.
In our work, the 2D states by themselves do not introduce any attention mechanism. 
The visualization method is mainly for the purpose of interpretation,
which helps us better understand the internal
dynamics of the decoding process.
To our best knowledge, this is accomplished for the first
time for image captioning.


\section{Formulations}
\label{sec:frm}

To begin with, we review the encoder-decoder framework~\cite{vinyals2015show}
which represents latent states as 1D vectors.
Subsequently, we reformulate the latent states as multi-channel 2D feature maps
for this framework.
These formulations are the basis for our comparative study.

\subsection{Encoder-Decoder for Image Captioning}

The encoder-decoder framework generates a caption for a given image
in two stages, namely \emph{encoding} and \emph{decoding}.
Specifically, given an image $I$, it first encodes the image
into a feature vector $\vv$, with a \emph{Convolutional Neural Network (CNN)},
such as VGGNet~\cite{simonyan2014very} or ResNet~\cite{He2015}.
The feature vector $\vv$ is then fed to a \emph{Recurrent Neural Network (RNN)}
and decoded into a sequence of words $(w_1, \ldots, w_T)$.
For decoding, the RNN implements a recurrent process driven by latent states,
which generates the caption through multiple steps, each yielding a word.
Specifically, it maintains a set of latent states, represented by a vector
$\vh_t$ that would be updated along the way. The computational procedure
can be expressed by the formulas below:
\begin{align}
	& \vh_0 = \vzero, \quad
	\vh_t = g(\vh_{t - 1}, \vx_t, \mI), \\
	& \vp_{t|1:t-1} = \text{Softmax}(\mW_p \vh_t), \label{eq:pvec} \\
	& w_t \sim \vp_{t|1:t-1}. \label{eq:draw_w}
\end{align}
The procedure can be explained as follows.
First, the latent state $\vh_0$ is initialized to be zeros.
At the $t$-th step, $\vh_t$ is updated by an RNN cell $g$,
which takes three inputs:
the previous state $\vh_{t-1}$,
the word produced at the preceding step (represented by an
embedded vector $\vx_t$), and
the visual feature $\vv$.
Here, the cell function $g$ can take a simple form:
\begin{equation} \label{eq:rnn_update}
	g(\vh, \vx, \vv)
	= \tanh \left(\mW_h \vh + \mW_x \vx + \mW_v \vv\right).
\end{equation}
More sophisticated cells, such as GRU \cite{cho2014learning}~and LSTM \cite{hochreiter1997long},
are also increasingly adopted in practice.
To produce the word $w_t$, the latent state $\vh_t$ will be transformed
into a probability vector $\vp_{t|1:t-1}$ via a fully-connected
linear transform $\mW_p \vh_t$ followed by a softmax function.
Here, $\vp_{t|1:t-1}$ can be considered as the probabilities of $w_t$
conditioned on previous states.

Despite the differences in their architectures,
all existing RNN-based captioning models represent latent states as
\emph{vectors} without explicitly preserving the spatial structures.
In what follows, we will discuss the alternative choice that represents latent states as 2D multi-channel feature maps.

\subsection{From 1D to 2D}
\label{sec:1dto2d}

From a technical standpoint, a natural way to maintain spatial structures in latent states
is to formulate them as 2D maps and employ convolutions for state transitions,
which we refer to as RNN-2DS.

\begin{figure}[t]
	\centering
	\includegraphics[width=\textwidth]{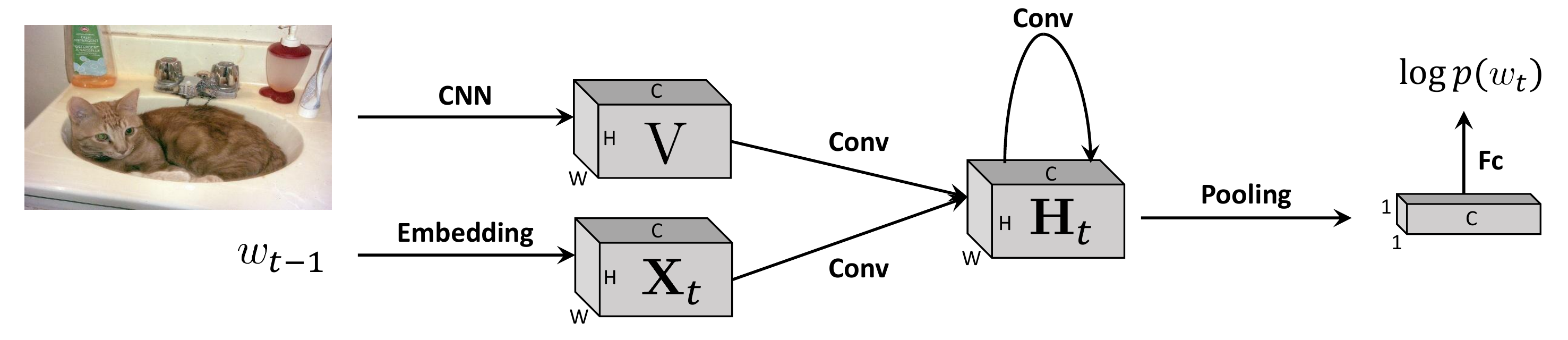}
	\caption{
		The overall structure of the encoder-decoder framework with RNN-2DS.
		Given an image $I$, a CNN first turns it into a multi-channel
		feature map $\mV$ that preserves high-level spatial structures.
		$\mV$ will then be fed to an RNN-2DS, where the latent state
		$\mH_t$ is also represented by multi-channel maps and the
		state transition is via convolution. At each step, the
		2D states are transformed into a 1D vectors and then decoded
		into conditional probabilities of words.
	}
	\label{fig:rnn2ds}
\end{figure}

Specifically, as shown in Figure \ref{fig:rnn2ds},
the visual feature $\mV$, the latent state $\mH_t$, and the word embedding
$\mX_t$ are all represented as 3D tensors of size $C \times H \times W$.
Such a tensor can be considered as a multi-channel map, which comprises
$C$ channels, each of size $H \times W$.
Unlike the normal setting
where the visual feature is derived from the activation of a fully-connected layer,
$\mV$ here is derived from the activation of a convolutional layer
that preserves spatial structures.
And $\mX_t$ is the 2D word embedding for $w_{t - 1}$,
of size $C \times H \times W$.
To reduce the number of parameters,
we use a lookup table of smaller size $C_x \times H_x \times W_x$
to fetch the raw word embedding, which will be enlarged to $C \times H \times W$ by two convolutional layers
\footnote{In our experiments, the raw word embedding is of size $4 \times 15 \times 15$, 
and is scaled up to match the size of latent states via 
two convolutional layers respectively with kernel sizes $32 \times 4 \times 5 \times 5$ and $C \times 32 \times 5 \times 5$.}.
With these representations, state updating can then be formulated using
\emph{convolutions}. For example, Eq.\eqref{eq:rnn_update} can be
converted into the following form:
\begin{equation}
	\mH_t = \text{relu} \left(
		\mK_h \conv \mH_{t-1} +
		\mK_x \conv \mX_t +
		\mK_v \conv \mV
	\right).
\end{equation}
Here, $\conv$ denotes the convolution operator,
and $\mK_h$, $\mK_x$, and $\mK_v$ are convolution kernels of size $C \times C \times H_k \times W_k$.
It is worth stressing that the modification presented above
is very flexible and can easily incorporate more sophisticated cells.
For example, the original updating formulas of GRU are
\begin{align}
	\vr_t & = \sigma(\mW_{rh} \vh_{t-1} + \mW_{rx} \vx_t + \mW_{rv} \vv), \notag \\
	\vz_t & = \sigma(\mW_{zh} \vh_{t - 1} + \mW_{zx} \vx_t + \mW_{zv} \vv), \notag \\
	\tilde \vh_t & = \tanh(\vr_t \circ (\mW_{hh} \vh_{t-1}) + \mW_{hx} \vx_t + \mW_{hv}\vv), \notag \\
	\vh_t & = \vz_t \circ \vh_{t - 1} + (1 - \vz_t) \circ \tilde \vh_t,
\end{align}
where $\sigma$ is the sigmoid function,
and $\circ$ is the element-wise multiplication operator.
In a similar way, we can convert them to the 2D form as
\begin{align} \label{eq:gru_update}
	\mR_t & = \sigma(\mK_{rh} \conv \mH_{t - 1} + \mK_{rx} \conv \mX_t + \mK_{rv} \conv \mV), \notag \\
	\mZ_t & = \sigma(\mK_{zh} \conv \mH_{t - 1} + \mK_{zx} \conv \mX_t + \mK_{zv} \conv \mV), \notag \\
	\tilde \mH_t & = \text{relu}(\mR_t \circ (\mK_{hh} \conv \mH_{t - 1}) + \mK_{hx} \conv \mX_t + \mK_{hv} \conv \mV), \notag \\
	\mH_t & = \mZ_t \circ \mH_{t - 1} + (1 - \mZ_t) \circ \tilde \mH_t.
\end{align}

Given the latent states $\mH_t$, the word $w_t$ can be generated
as follows. First, we compress $\mH_t$ (of size $C \times H \times W$)
into a $C$-dimensional vector $\vh_t$ by mean pooling
across spatial dimensions.
Then, we transform $\vh_t$ into a probability vector $\vp_{t|1:t-1}$ and
draw $w_t$ therefrom, following Eq.\eqref{eq:pvec} and \eqref{eq:draw_w}.
Note that the pooling operation could be replaced with more sophisticated modules,
such as an attention module,
to summarize the information from all locations for word prediction.
We choose the pooling operation as it adds zero extra parameters,
which makes the comparison between 1D and 2D states fair.

Since this reformulation is generic, besides the encoder-decoder framework,
it can be readily extended to other captioning models that adopt
RNNs as the language module, \eg~Att2in \cite{rennie2016self} and Review Net \cite{yang2016review}.


\section{Qualitative Studies on 2D States}
\label{sec:qualrst}

Thanks to the preserved spatial locality, the use of 2D states makes the
framework amenable to some qualitative analysis. Taking advantage of this,
we present three studies in this section:
(1) We manipulate the 2D states
and investigate how it impacts the generated captions. The results of this
study would corroborate the statement that 2D states help to preserve
spatial structures.
(2) Leveraging the spatial locality, we identify the associations
between the activations of latent states and certain subregions of the input
image.
Based on the dynamic associations between state activations and the corresponding subregions,
we can visually reveal the internal dynamics of the decoding process.
(3) Through latent states we also interpret the connections between the visual
and the linguistic domains.

\subsection{State Manipulation}

\begin{figure}[t]
\centering
\includegraphics[width=\textwidth]{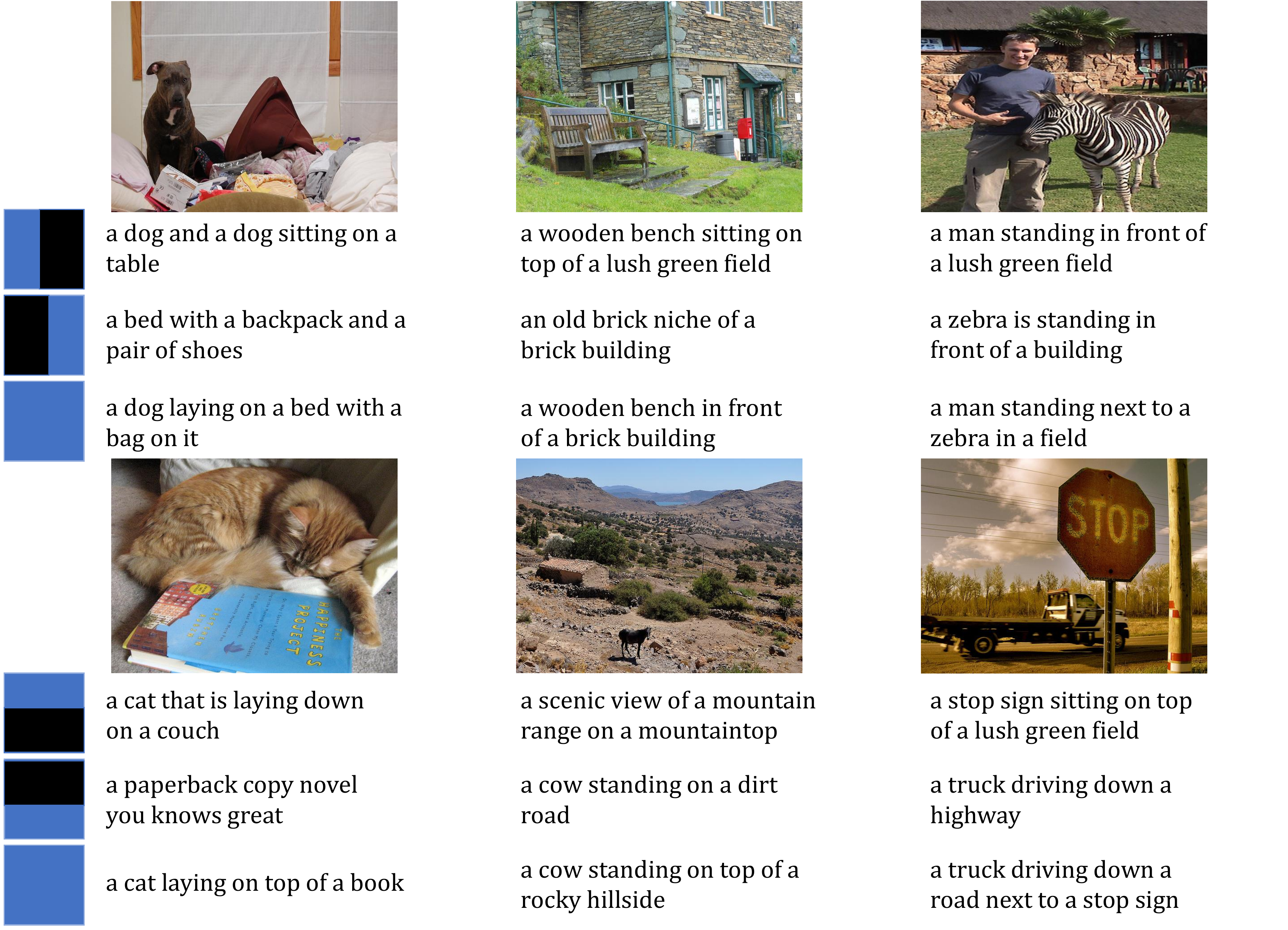}
\caption{This figure lists several images with generated captions relying on various parts of RNN-2DS's states.
The accessible part is marked with \textcolor{blue}{blue} color in each case.}
\label{fig:manipulation}
\end{figure}

We study how the spatial structures of the 2D
latent states influence the resultant captions by controlling the
accessible parts of the latent states.

As discussed in Sec.~\ref{sec:1dto2d},
the prediction at $t$-th step is based on $\vh_t$,
which is pooled from $\mH_t$ across $H$ and $W$.
In other words, $\vh_t$
summarizes the information from the entire area of $\mH_t$.
In this experiment,
we replace the original region $(1, 1, H, W)$ with a subregion between
the corners $(x_1, y_1)$ and $(x_2, y_2)$ to get a modified summarizing vector $\vh_t^\prime$ as
\begin{align}
	\vh_t^\prime = \frac{1}{(y_2 - y_1 + 1) (x_2 - x_1 + 1)}\sum_{i = y_1}^{y_2} \sum_{j = x_1}^{x_2} \mH_t |_{(i, j)}.
\end{align}
Here, $\vh_t^\prime$ only captures a subregion of the image, on which
the probabilities for the word $w_t$ is computed.
We expect that this caption only partially reflects the visual semantics.

Figure~\ref{fig:manipulation} shows several images together with
the captions generated using different subregions of the 2D states.
Take the bottom-left image in Figure~\ref{fig:manipulation} for an instance,
when using only the upper half of the latent states,
the decoder generates a caption focusing on the cat,
which indeed appears in the upper half of the image.
Similarly,
using only the lower half of the latent states results in a caption that
talks about the book located in the lower half of the image.
In other words,
depending on a specific subregion of the latent states,
a decoder with 2D states tends to generate a caption that conveys the visual content
of the corresponding area in the input image.
This observation suggests that the 2D latent states do preserve the spatial
structures of the input image.

Manipulating latent states differs essentially from
the passive data-driven attention module \cite{xu2015show} commonly adopted in
captioning models.
It is a controllable operation,
and does not require a specific module to achieve such functionality.
With this operation,
we can extend a captioning model with 2D states to allow
\emph{active} management of the focus, which, for example, can be
used to generate multiple complementary sentences for an image.
While the attention module can be considered as an automatic manipulation on latent states,
the combination of 2D states and the attention mechanism worths exploring in the future work. 

\subsection{Revealing Decoding Dynamics}
\label{sec:dynamics}

This study intends to analyze internal dynamics of the decoding process,
\ie~how the latent states evolve in a series of decoding steps.
We believe that it can help us better understand
how a caption is generated based on the visual content.
The spatial locality of the 2D states
allows us to study this in an efficient and effective way.

\begin{figure}[t]
	\centering
	\includegraphics[width=\textwidth]{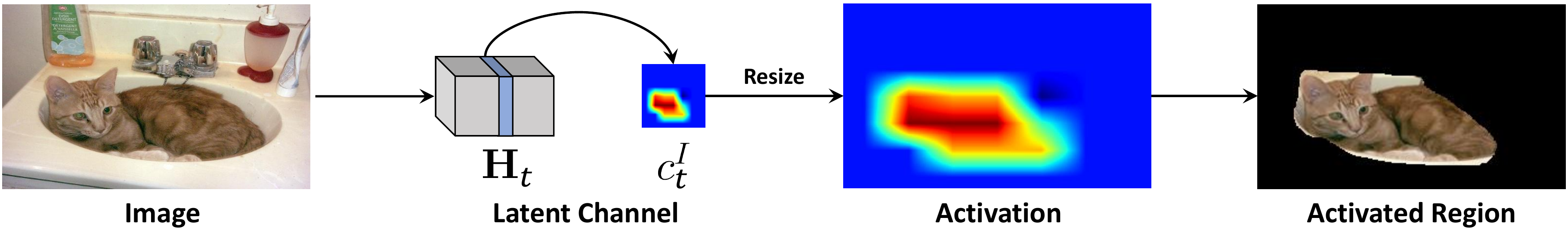}
	\caption{
	This figure shows our procedure of finding the activated region of a
	latent channel at the $t$-th step.
	}
	\label{fig:actreg}
\end{figure}

We use \emph{activated regions} to align the activations of the latent states at different decoding steps
with the subregions in the input image.
Specifically, we treat the channels of 2D states as the basic units in our study,
which are 2D maps of activation values.
Given a state channel $c$ at the $t$-th decoding step,
we resize it to the size of the input image $I$ via bicubic interpolation.
The pixel locations in $I$ whose corresponding interpolated activations are above a certain
threshold \footnote{See released code for more details.}
are considered to be \emph{activated}.
The collection of all such pixel locations is referred to as the \emph{activated region}
for the state channel $c$ at the $t$-th decoding step,
as shown in Figure~\ref{fig:actreg}.

\begin{figure}[t]
\centering
\includegraphics[width=\textwidth]{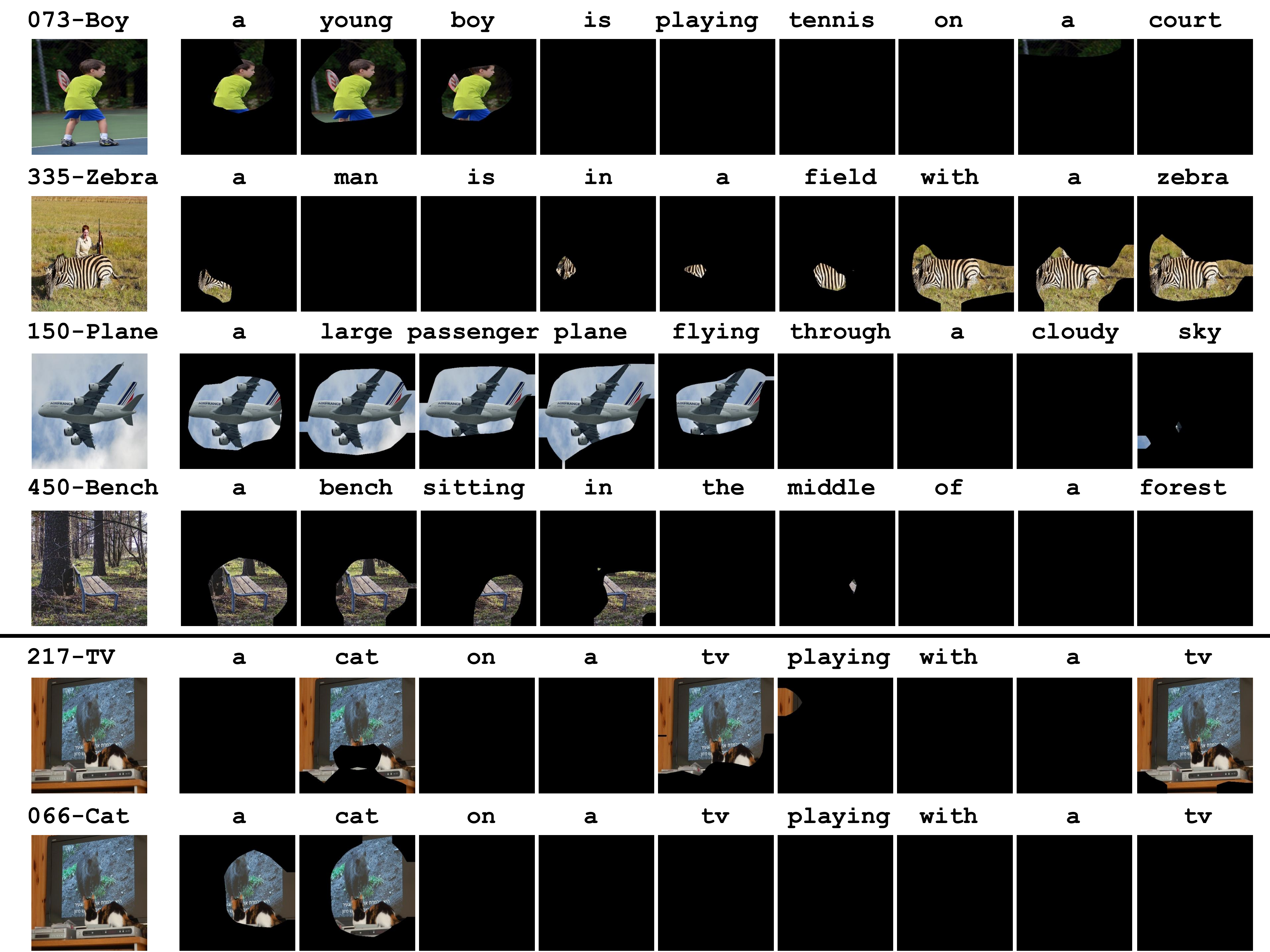}
\caption{This figure shows the changes of several channels, in terms of the activated regions, during the decoding processes.
 On the last two cases, changes of two channels in the same decoding process are shown and compared.
 (Best viewed in high resolution)}
\label{fig:seqchannel}
\end{figure}

With activated regions computed respectively at different decoding steps for one state channel,
we may visually reveal the internal dynamics of the decoding process at that channel.
Figure~\ref{fig:seqchannel} shows several images and their generated captions,
along with the activated regions of some channels following the decoding processes.
These channels are selected as they are associated with nouns in the generated captions,
which we will introduce in the next section.
Via this study we found that
(1) The activated regions of channels often capture salient visual entities in the image,
and also reflect the surrounding context occasionally.
(2) During a decoding process,
different channels have different dynamics.
For a channel associated with a noun,
the activated regions of its associated channel become significant as the decoding process approaches
the point where the noun is produced,
and the channel becomes deactivated afterwards.

The revealed dynamics can help us better understand the decoding process,
which also point out some directions for future study.
For instance,
in Figure \ref{fig:seqchannel},
the visual semantics are distributed to different channels,
and the decoder moves its focus from one channel to another.
The mechanism that triggers such movements remains needed to be explored.

\subsection{Connecting Visual and Linguistic Domains}
\label{sec:association}

\begin{figure}[t]
\centering
\includegraphics[width=\textwidth]{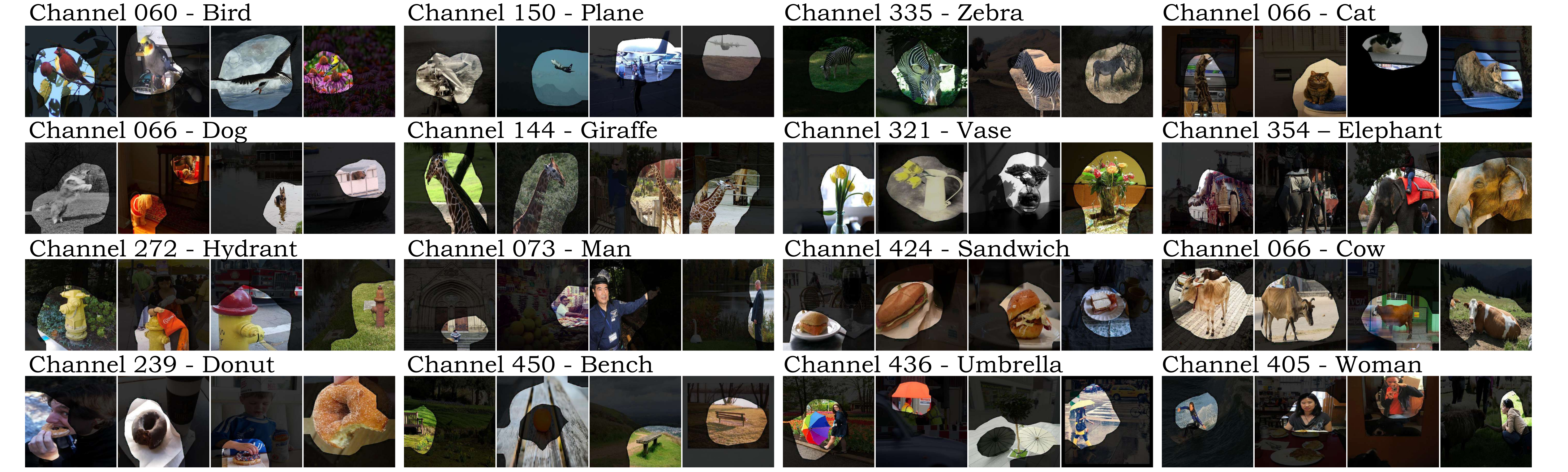}
\caption{
Sample words and their associated channels in \emph{RNN-2DS-($512, 7, 7$)}.
For each word, $5$ activated regions of its associated channel on images that contain this word in the generated captions are shown.
The activated regions are chosen at the steps where the words are produced. (Best viewed in high resolution)}
\label{fig:wordchannel}
\end{figure}

Here we investigate how the visual domain is connected to the linguistic domain.
As the latent states serve as pivots that connect both domains,
we try to use the activations of the latent states to identify the detailed
connections.

First, we find the associations between the latent states and the words.
Similar to Sec.~\ref{sec:dynamics},
we use state channels as the basic units here,
so that we can use the activated regions which connect the
latent states to the input image.
In Sec.~\ref{sec:dynamics}, we have observed that
a channel associated with a certain word is likely to remain active until the word is produced,
and its activation level will drop significantly afterwards thus preventing that word from being generated again.
Hence, one way to judge whether a channel is associated with a word
is to estimate the difference in its level of activations before and after the word is generated.
The channel that yields \emph{maximum difference} can be considered as the one associated
with the word \footnote{See released code for more details.}.

\textbf{Words and Associated Channels.}
For each word in the vocabulary, we could find its associated channel
as described above,
and study the corresponding activated regions,
as shown in Figure~\ref{fig:wordchannel}.
We found that
(1) Only nouns have strong associations with the state channels, which is
consistent with the fact that spatial locality is highly-related with
the visual entities described as nouns.
(2) Some channels have multiple associated nouns. For example,
\emph{Channel}-$066$ is associated with
\emph{``cat''}, \emph{``dog''}, and \emph{``cow''}.
This is not surprising -- since there are more nouns in the vocabulary than
the number of channels, some nouns have to share channels.
Here, it is worth noting that the nouns that share a channel tend to be
visually relevant. This shows that the latent channels
can capture meaningful visual structures.
(3) Not all channels have associated words. Some channels may capture
abstract notions instead of visual elements. The study of such
channels is an interesting direction in the future.

\textbf{Match of Words and Associated Channels.}
On top of the activated regions, we could also estimate the match between a word and its associated channel.
Specifically,
noticing the activated regions visually look like the attention maps in \cite{liu2017attention},
we borrow the measurement of attention correctness from \cite{liu2017attention},
to estimate the match.
\emph{Attention correctness} computes the similarity between a human-annotated segmentation mask of a word,
and the activated region of its associated channel, at the step the word is produced.
The computation is done by summing up the normalized activations within that mask.
On MSCOCO~\cite{lin2014microsoft}, we evaluated the attention correctness on $80$ nouns that have human-annotated masks.
As a result, the averaged attention correctness is $0.316$.
For reference, following the same setting except for replacing the activated regions with the attention maps,
AdaptiveAttention~\cite{lu2016knowing}, a state-of-the-art captioning model,
got a result of $0.213$.

\begin{figure}[t]
\centering
\includegraphics[width=\textwidth]{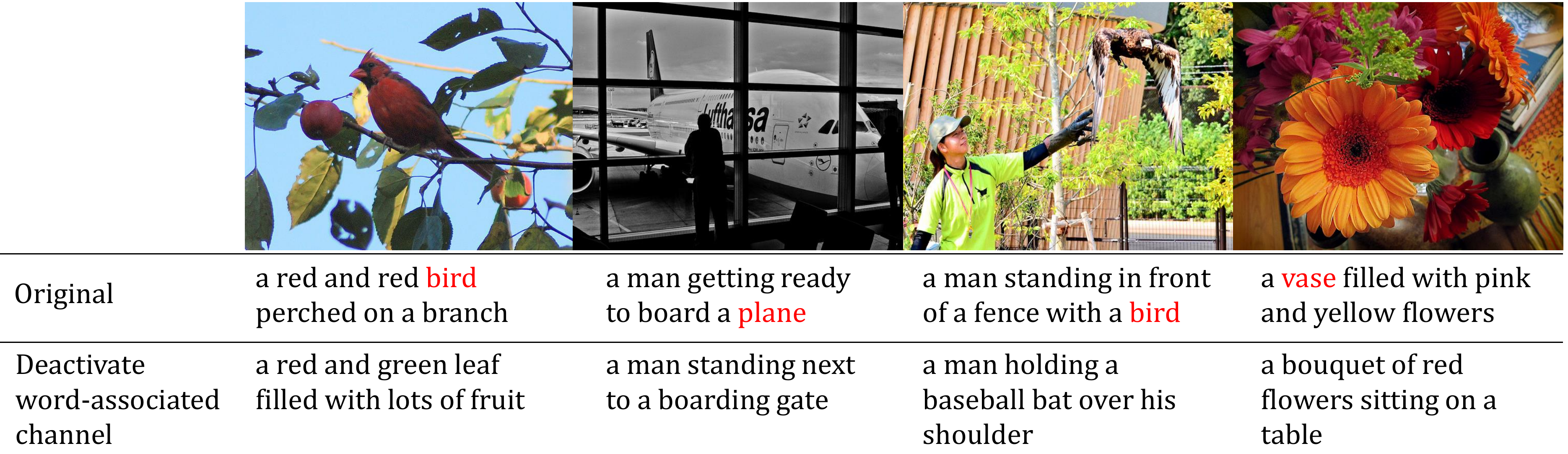}
\caption{This figure lists some images with generated captions before and after some word-associated channel being deactivated.
The word that associates with the deactivated channel is marked in \textcolor{red}{red}.}
\label{fig:channelablation}
\end{figure}

\textbf{Deactivation of Word-Associated Channels.}
We also verify the match of the found associations between the state channels and the words alternatively via an ablation study,
where we compare the generated captions with and without the involvement of a certain channel.
Specifically,
on images that contain the target word $w$ in the generated captions,
we re-run the decoding process,
in which we deactivate the associated channel of $w$ by clipping its value to zero at all steps,
then compare the generated captions with previous ones.
As shown in Figure \ref{fig:channelablation},
deactivating a word-associated channel leads to the miss of the corresponding words in the generated captions,
even though the input still contains the visual semantics for those words.
This ablation study corroborates the validity of our found associations.


\section{Comparison on Captioning Performance}
\label{sec:quanrst}

In this section, we compare the encoder-decoder framework
with 1D states and 2D states.
Specifically,
we run our studies on MSCOCO~\cite{lin2014microsoft} and Flickr30k~\cite{young2014image},
where we at first introduce the settings, followed by the results.

\subsection{Settings}

MSCOCO~\cite{lin2014microsoft} contains $122,585$ images.
We follow the splits in \cite{karpathy2015deep},
using $112,585$ images for training, $5,000$ for validation, and the
remaining $5,000$ for testing.
Flickr30K~\cite{young2014image}~contains $31,783$ images in total,
and we follow splits in \cite{karpathy2015deep},
which has $1,000$ images respectively for validation and testing,
and the rest for training.
In both datasets, each image comes with $5$ ground-truth captions.
To obtain a vocabulary, we turn words to lowercase and remove those with non-alphabet characters.
Then we replace words
that appear less than $6$ times with a special token \emph{UNK},
resulting in a vocabulary of size $9,487$ for MSCOCO, and $7,000$ for Flickr30k.
Following the common convention~\cite{karpathy2015deep}, we truncated all ground-truth
captions to have at most $18$ words.

All captioning methods in our experiments are based on the encoder-decoder
paradigm~\cite{vinyals2015show}. We use
ResNet-152~\cite{He2015} pretrained on ImageNet \cite{ILSVRC15} as the encoder
in all methods.
In particular, we take the output of the layer \texttt{res5c} as
the visual feature $\mV$.
We use the combination of the cell type and the state shape to refer to each type of the decoder.
\eg~\emph{LSTM-1DS-($L$)} refers to a standard LSTM-based decoder with latent states of size $L$,
and \emph{GRU-2DS-($C,H,W$)} refers to an RNN-2DS decoder with GRU cells as
in Eq.\eqref{eq:gru_update}, whose latent states are of size $C \times H \times W$.
Moreover, all RNN-2DS models adopt a raw word-embedding of size
$4 \times 15 \times 15$, except when a different size is explicitly specified.
The convolution kernels $\mK_h$, $\mK_x$, and $\mK_v$ share the same size
$C \times C \times 3 \times 3$. 

The focus of this paper is the representations of latent states.
To ensure fair comparison,
no additional modules including the attention module \cite{xu2015show}~are added to the methods.
Moreover, no other training strategies are utilized, such as the scheduled sampling \cite{bengio2015scheduled},
except for the maximum likelihood objective,
where we use the ADAM optimizer~\cite{kingma2014adam}.
During training, we first fix the CNN encoder and optimize the decoder
with learning rate $0.0004$ in the first $20$ epochs, and then
jointly optimize both the encoder and decoder,
until the performance on the validation set saturates.

For evaluation, we report the results using metrics including
BLEU-4 (B4)~\cite{papineni2002bleu}, METEOR (MT)~\cite{lavie2014meteor},
ROUGE (RG)~\cite{lin2004rouge}, CIDER (CD)~\cite{vedantam2015cider},
and SPICE (SP)~\cite{anderson2016spice}.

\subsection{Comparative Results}

\begin{figure}[t]
\centering
\includegraphics[width=\textwidth]{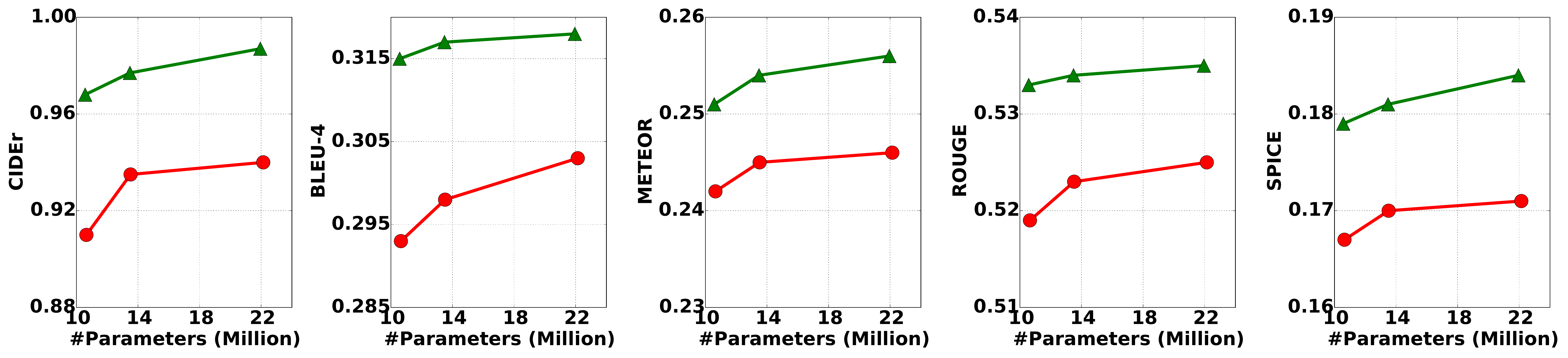}
\caption{The results, in terms of different metrics, 
obtained using RNN-2DS (\textcolor{green}{green}) and LSTM-1DS (\textcolor{red}{red}) 
on the MSCOCO offline test set with similar parameter sizes. 
Specifically, RNN-2DS of sizes 10.57M, 13.48M and 21.95M have compared to LSTM-1DS of sizes 10.65M, 13.52M and 22.14M.} 
\label{fig:curve_1dvs2d}
\end{figure}

First, we compared \emph{RNN-2DS} with \emph{LSTM-1DS}.
The former has 2D states with the simplest type of cells
while the latter has 1D states with sophisticated LSTM cells.
As the capacity of a model is closely related to the number of parameters,
to ensure a fair comparison, each config of \emph{RNN-2DS} is compared to
an \emph{LSTM-1DS} config \emph{with a similar number of parameters}.
In this way, the comparative results will signify the differences in
the inherent expressive power of both formulations.

The resulting curves in terms of different metrics are shown in Figure \ref{fig:curve_1dvs2d},
in which we can see that
\emph{RNN-2DS} outperforms
\emph{LSTM-1DS} consistently,
across different parameter sizes and under different metrics.
These results show that \emph{RNN-2DS}, with the states that preserve
spatial locality, can capture both visual and linguistic information
more efficiently.

\begin{table}[t]
\centering
\caption{
	The results obtained using different decoders on the offline and online test sets of MSCOCO, and the test set of Flickr30k, where METEOR (MT) \cite{lavie2014meteor} is omitted due to space limitation, and no SPICE (SP) \cite{anderson2016spice} is reported by the online test set of MSCOCO.
}
\scriptsize
\begin{tabular}{cc ccccccc c ccccc c ccccccc}
\toprule
	\multirow{2}{*}{Model} & \multirow{2}{*}{\#Param} & \multicolumn{7}{c}{COCO-offline} && \multicolumn{5}{c}{COCO-online} && \multicolumn{7}{c}{Flickr30k} \\
	\cmidrule{3-9} \cmidrule{11-15} \cmidrule{17-23}
		& & CD && B4 && RG && SP & & CD && B4 && RG & & CD && B4 && RG && SP \\
\midrule
	RNN-1DS-(595)	& 13.58M & 0.914 && 0.293 && 0.520 && 0.168 & & 0.868 && 0.286 && 0.515 & & 0.353 && 0.195 && 0.427 && 0.117 \\
	GRU-1DS-(525)	& 13.53M & 0.920 && 0.295 && 0.520 && 0.169 & & 0.889 && 0.291 && 0.518 & & 0.360 && 0.195 && 0.428 && 0.117 \\ 
	LSTM-1DS-(500)	& 13.52M & 0.935 && 0.298 && 0.523 && 0.170 & & 0.904 && 0.295 && 0.523 & & 0.381 && 0.202 && 0.437 && 0.120 \\
	\midrule
	RNN-2DS-(256,7,7) & 13.48M & 0.977 && 0.317 && 0.534 && 0.181 & & 0.930 && 0.305 && 0.527 & & 0.420 && 0.217 && 0.442 && 0.125 \\
	GRU-2DS-(256,7,7) & 17.02M & 1.001 && 0.323 && 0.539 && 0.186 & & 0.962 && 0.316 && 0.535 & & 0.438 && 0.218 && 0.445 && 0.131 \\
	LSTM-2DS-(256,7,7) & 18.79M & 0.994 && 0.319 && 0.538 && 0.187 & & 0.958 && 0.313 && 0.531 & & 0.427 && 0.220 && 0.444 && 0.132 \\ 
\bottomrule
\end{tabular}
\label{tab:rnn2ds_cell}
\end{table}

We also compared different types of decoders with similar numbers of parameters,
namely \emph{RNN-1DS}, \emph{GRU-1DS}, \emph{LSTM-1DS}, \emph{RNN-2DS}, \emph{GRU-2DS}, and \emph{LSTM-2DS}.
Table \ref{tab:rnn2ds_cell} shows the results of these decoders on both datasets,
from which we observe:
(1) \emph{RNN-2DS} outperforms \emph{RNN-1DS}, \emph{GRU-1DS}, and \emph{LSTM-1DS},
indicating that embedding latent states in 2D forms is more effective.
(2) \emph{GRU-2DS}, which is also based on the proposed formulation but adds several gate functions,
surpasses other decoders and yields the best result.
This suggests that
the techniques developed for conventional RNNs including gate functions
and attention modules~\cite{xu2015show} are very likely to benefit RNNs
with 2D states as well.

\begin{figure}[t]
\centering
\includegraphics[width=\textwidth]{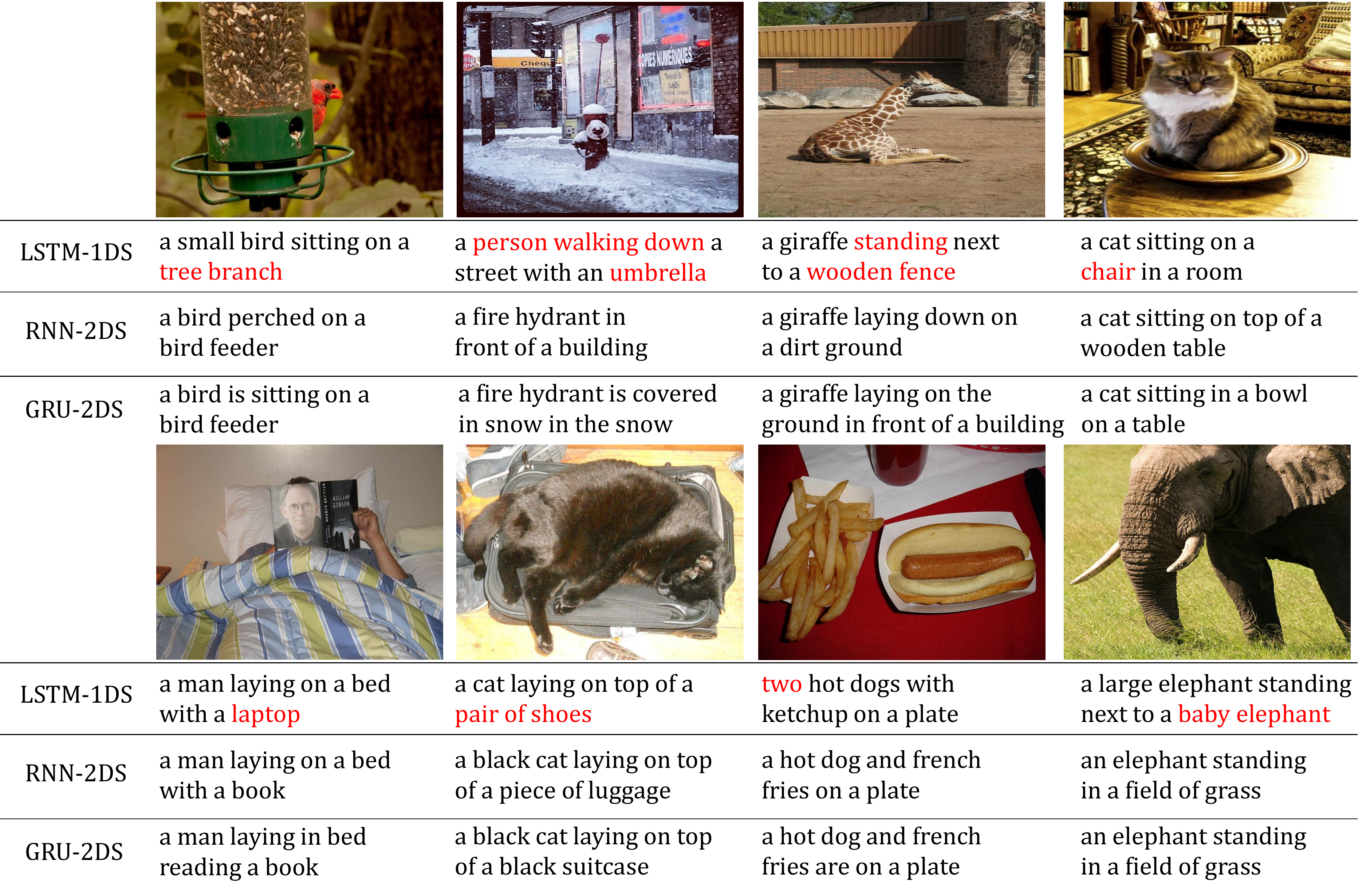}
\caption{This figure shows some qualitative samples of captions generated by different decoders,
where words in \textcolor{red}{red} indicate they are inconsistent with the image.}
\label{fig:qualitative}
\end{figure}

Figure~\ref{fig:qualitative} includes some qualitative samples,
in which we can see
the captions generated by \emph{LSTM-1DS} rely heavily on the language priors,
which sometimes contain the phrases that are not consistent with the visual
content but appear frequently in training captions.
On the contrary, the sentences from \emph{RNN-2DS} and \emph{GRU-2DS}
are more relevant to the visual content.

\subsection{Ablation Study}

\begin{table}[t]
\centering
\caption{
	The results obtained on the MSCOCO offline test set using RNN-2DS with different choices on 
	pooling functions, activation functions, word-embeddings, kernels and latent states.
	Except for the first row, each row only lists the choice that is different from the first row. 
	"-" means the same.  
}
\scriptsize
\begin{tabular}{cccccccccc cc cc cc cc cc}
\toprule
	Pooling && Activation && Word-Embedding && Kernel && Latent-State && CD && B4 && MT && RG && SP  \\
\midrule
	 Mean && ReLU && $4\times15\times15$ && $3\times3$ && $256\times7\times7$ && 0.977 && 0.317 && 0.254 && 0.534 && 0.181 \\
\midrule
	 - && tanh && - && - && - && 0.924 && 0.302 && 0.244 && 0.522 && 0.174 \\
	 Max && - && - && - && - && 0.850 && 0.279 && 0.233 && 0.507 && 0.166 \\
\midrule
 	 - && - && $1\times15\times15$ && - && - && 0.965 && 0.313 && 0.251 && 0.532 && 0.180 \\
	 - && - && $7\times15\times15$ && - && - && 0.951 && 0.309 && 0.250 && 0.529 && 0.179 \\
\midrule
	 - && - && - && $1\times1$ && - && 0.927 && 0.298 && 0.247 && 0.522 && 0.177 \\
	 - && - && - && $5\times5$ && - && 0.951 && 0.308 && 0.250 && 0.529 && 0.177 \\
\midrule
	 - && - && - && - && $256\times5\times5$ && 0.934 && 0.300 && 0.245 && 0.523 && 0.173 \\
	 - && - && - && - && $256\times11\times11$ && 0.927 && 0.300 && 0.246 && 0.523 && 0.176 \\ 	
\bottomrule
\end{tabular}
\label{tab:designchoice}
\end{table}

Table \ref{tab:designchoice} compares the performances obtained with
different design choices in \emph{RNN-2DS},
including pooling methods, activation functions, and sizes of word embeddings, kernels and latent states
The results show that mean pooling outperforms
max pooling by a significant margin, indicating that information from all
locations is significant.
The table also shows the best combination of modeling choices for RNN-2DS:
mean pooling, ReLU, the word embeddings of size $4\times15\times15$,
the kernel of size $3\times3$, and the latent states of size $256\times7\times7$.


\section{Conclusions and Future Work}
\label{sec:concls}

In this paper, we studied the impact of embedding latent states as 2D multi-channel feature maps
in the context of image captioning.
Compared to the standard practice that embeds latent states as 1D vectors,
2D states consistently achieve higher captioning performances across different settings.
Such representations also preserve the spatial locality of the latent states,
which helps reveal the internal dynamics of the decoding process,
and interpret the connections between visual and linguistic domains.
We plan to combine the decoder having 2D states with 
other modules commonly used in captioning community,
including the attention module \cite{xu2015show},
for further exploration.

\paragraph{Acknowledgement}
This work is partially supported by the Big Data Collaboration Research grant from SenseTime Group (CUHK Agreement No. TS1610626), the General Research Fund (GRF) of Hong Kong (No. 14236516).

{
\bibliographystyle{splncs}
\bibliography{2dcaption}
}

\renewcommand\thesection{\Alph{section}}
\renewcommand\thesubsection{\thesection.\arabic{subsection}}
\setcounter{section}{0}

\newpage

\section{Combining 2D Models with Popular Techniques}
We add several popular techniques for image captioning models to the captioning model with 2D states,
and the results are listed in Table \ref{tab:tech}. 
We can see that when used in conjunction with reinforcement learning (RL), schedule sampling (SS) or an attention sub-net (ATT),
RNN-2DS is competitive against the state-of-the-arts.
This also suggests that new methods can also benefit from adopting the 2D state representation.

\begin{table}
\centering
\caption{Results on MSCOCO offline test set, obtained by applying RNN-2DS with several additional techniques,
as well as several baselines.}
\begin{tabular}{c|c|c|c|c|c|c|c|c}
\toprule
		& ERD[1] & MSM[2] & AdapAtt[3] & 1DS + ATT & 2DS & 2DS + SS & 2DS + RL & 2DS + ATT  \\
\midrule
	CIDEr		& 0.895 & 0.986 & 1.085 & 0.958 & 0.977 & 1.004 & 1.087 & 0.998 \\
\bottomrule
\end{tabular}
\label{tab:tech}
\end{table}

\section{Activated Regions}
For a given image $I$, the channel $c$ of $\mH_t$, denoted by $c_t^I$,
is a map of size $H \times W$.
To obtain the activated region, we first resize $c_t^I$ to the size of $I$
with bicubic interpolation, and then identify the activated pixels
by thresholding.
In particular, those pixels whose corresponding values in $c_t^I$ are above
the threshold $\lambda \cdot v^\star$ are considered as \emph{activated}.
Here, $v^\star$ is the maximum value in the corresponding channel $c^I_t$
over all decoding steps, and $\lambda$ is a coefficient in $[0, 1]$
that controls the range of the activated regions.
In practice, we set $\lambda = 0.2$.

\section{Word-Channel Association}

To identify the connections between latent states and words,
we devise a metric to measure the degree of \emph{association} between
a word $w$ and a channel $c$, denoted by $s(w, c)$.

The metric is designed following this observation:
a channel associated with a certain word is likely to remain active
until the word is produced, and its activation level will drop significantly
afterwards thus preventing that word from being generated again.

First, we measure the \emph{activation level} of a channel $c$ at the $t$-th step
on a given image $I$ as the sum of its entries:
\begin{equation} \label{eq:vis_s}
	\eta(c^I_t) = \sum_{i = 1}^H \sum_{j = 1}^W c^I_t(i, j).
\end{equation}
Then we use $A_{t_1, t_2}$ to denote the activation level averaged
over a certain period $[t_1, t_2]$, as
\begin{equation} \label{eq:vis_a}
	A_{t_1, t_2}(c^I) = \frac{1}{t_2 - t_1 + 1} \sum_{j = t_1}^{t_2} \eta(c^I_j).
\end{equation}
Finally, the \emph{association score} $s(w, c)$ is defined to be the difference
between the average activation level up to the step where $w$ is produced
and the average level afterwards. Such a difference is averaged over
all samples that contain the word $w$. Formally, this can be expressed as:
\begin{equation} \label{eq:vis_cor}
	s(w, c) = \frac{1}{|\cI(w)|}\sum_{I \in \cI(w)}
	A_{1, t^w_I}(c^I) - A_{t^w_I + 1,T_I}(c^I).
\end{equation}
Here,
$\cI(w)$ is the set of all images that contain $w$ in their \emph{generated} captions.
$T_I$ is the length of the caption for $I$.
$t^w_I$ is the step at which $w$ is produced for $I$.
$A_{1, t^w_I}(c^I)$ and $A_{t^w_I + 1, T_I}(c^I)$ are respectively
the average activations \emph{before} and \emph{after} $w$ is produced.
Based on the \emph{association score},
for each word $w$, we could find the \emph{most relevant} channel as
$c^\star = \argmax_{c} s(w, c)$.

\section{More Qualitative Results}

In this section we provide more qualitative results.
Specifically,
in Figure \ref{fig:manipulation} we show more images with captions generated by controlling the accessible parts in the latent states.
And more samples indicating the internal dynamics of the decoding process are shown in Figure \ref{fig:seqchannel}.
Figure \ref{fig:wordchannel} shows more words with their associated channels,
and Figure \ref{fig:channelablation} shows more images with generated captions with and without deactivating some word-associated channels.
Finally, in Figure \ref{fig:qualitative} we show more qualitative samples with captions generated by different decoders.

\begin{figure}[!htbp]
\centering
\includegraphics[width=\textwidth]{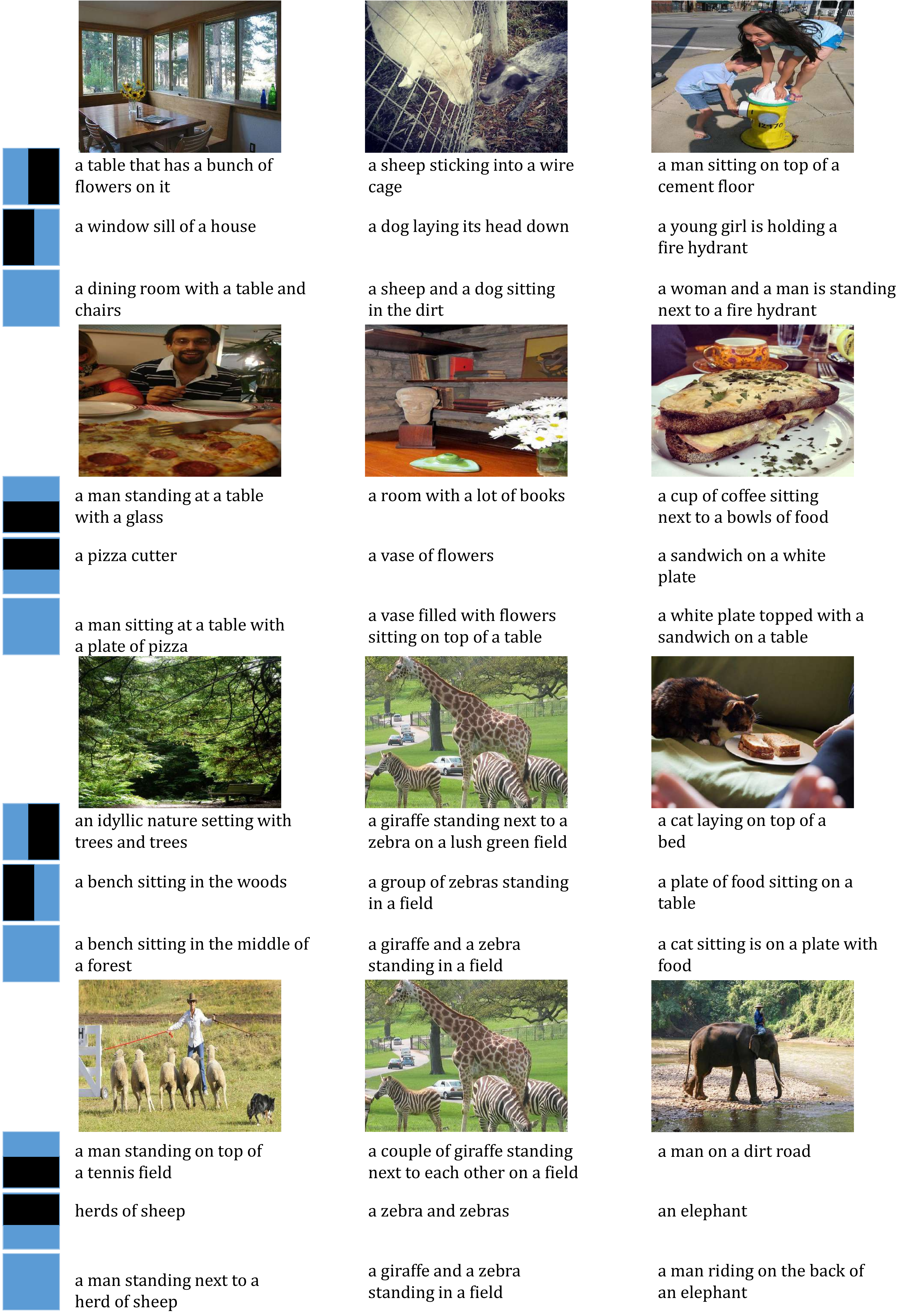}
\caption{More images with captions generated by controlling the accessible parts in the latent states.}
\label{fig:manipulation}
\end{figure}
\begin{figure}[!htbp]
\centering
\includegraphics[height=\textheight]{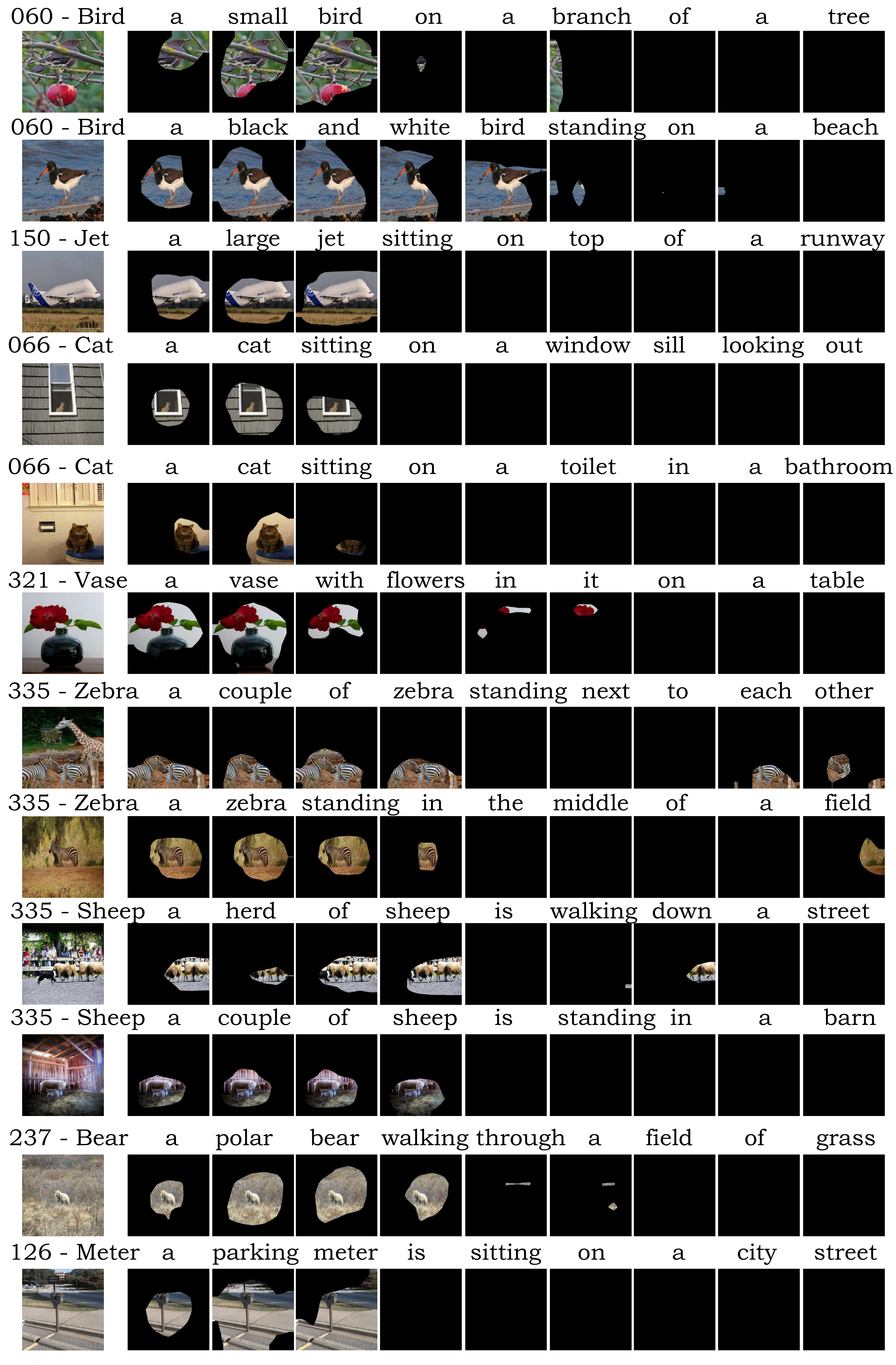}
\caption{More samples showing the changes of channels during the decoding processes.}
\label{fig:seqchannel}
\end{figure}
\begin{figure}[!htbp]
\centering
\includegraphics[width=\textwidth]{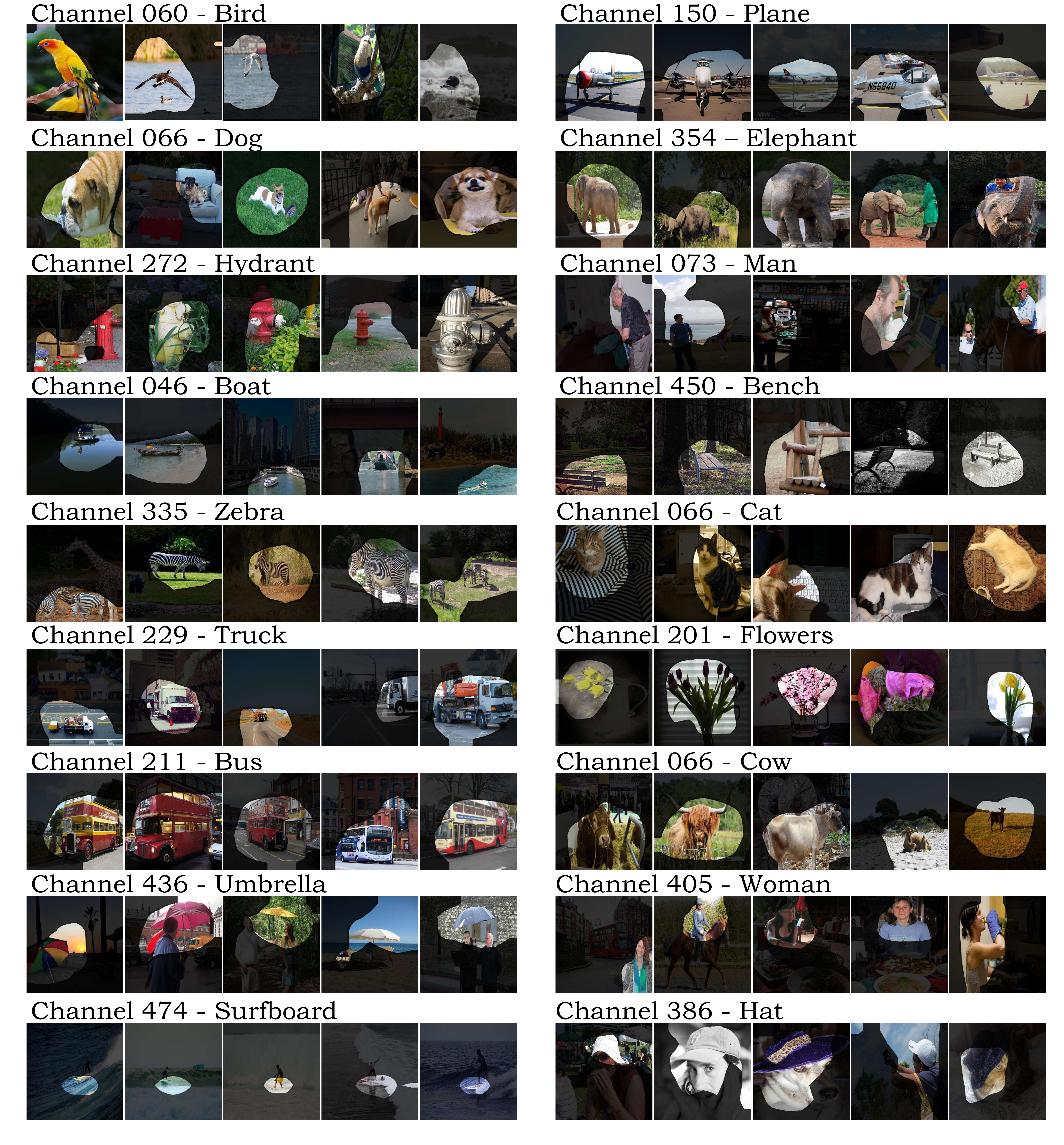}
\caption{More samples showing activated regions of channels and their associated words.}
\label{fig:wordchannel}
\end{figure}
\begin{figure}[!htbp]
\centering
\includegraphics[width=\textwidth]{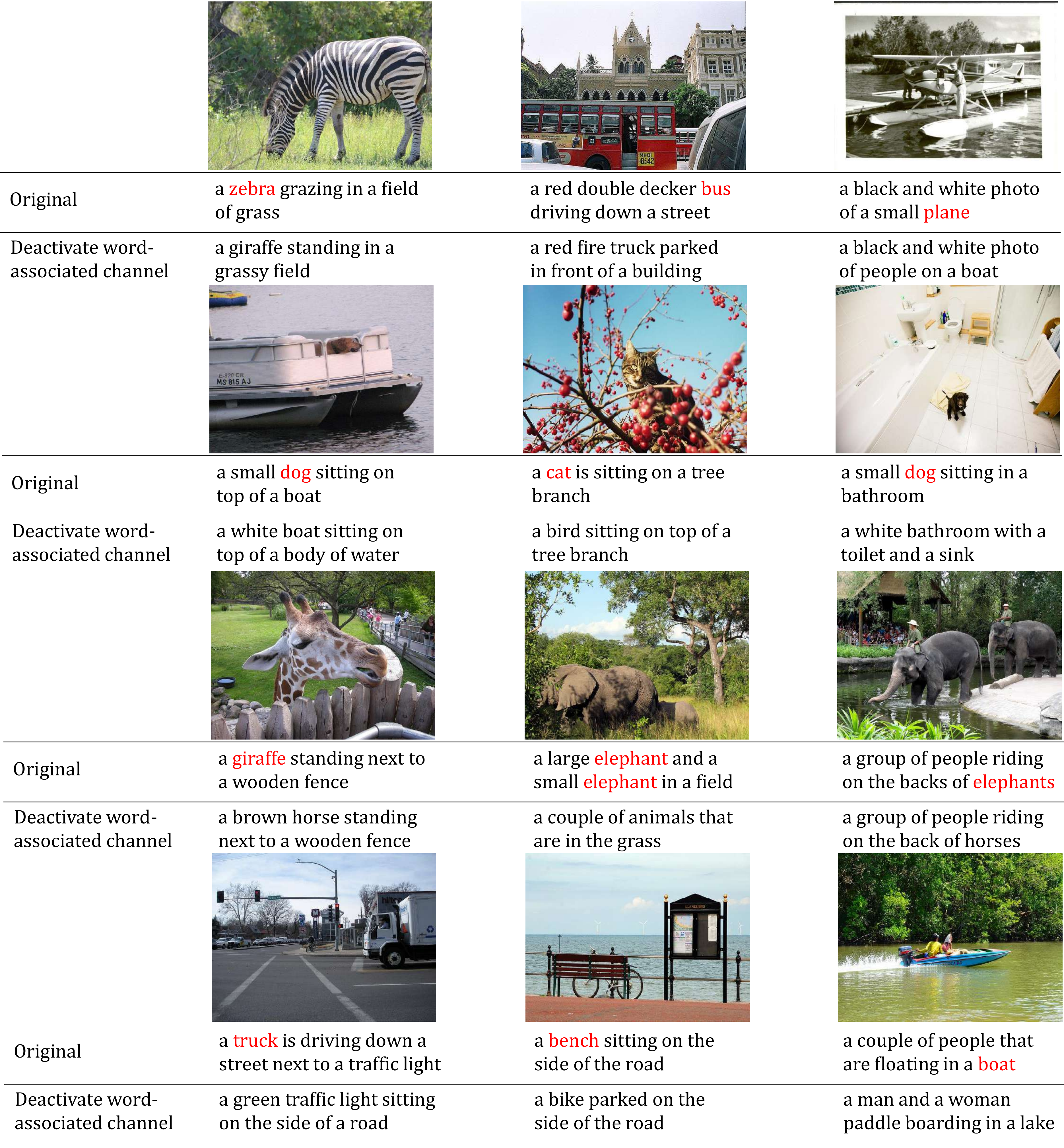}
\caption{More images with captions generated with and without some word-associated channels being deactivated.}
\label{fig:channelablation}
\end{figure}
\begin{figure}[!htbp]
\centering
\includegraphics[width=\textwidth]{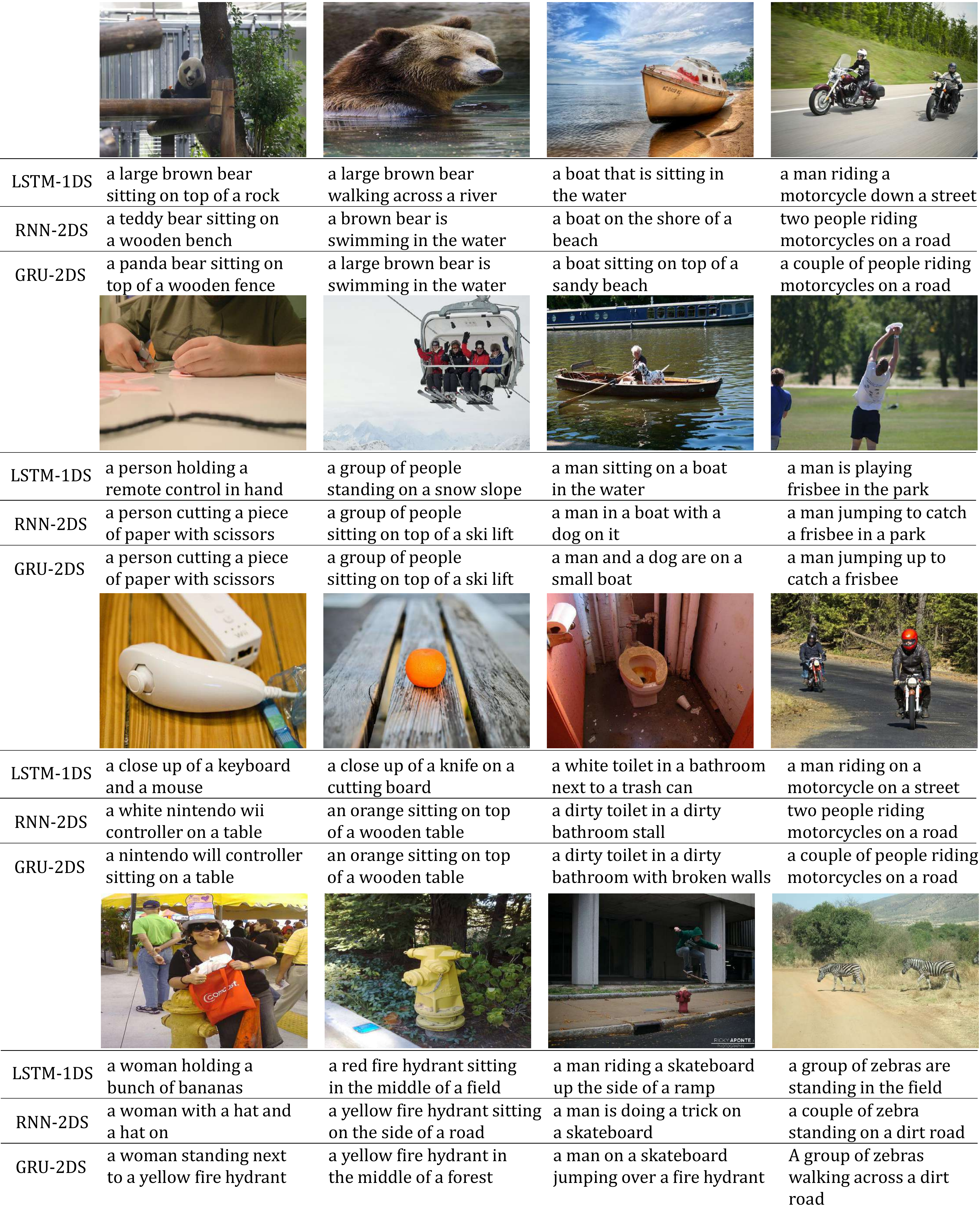}
\caption{More samples showing images and captions generated by different decoders.}
\label{fig:qualitative}
\end{figure}

\end{document}